\documentclass[conference]{IEEEtran}
\IEEEoverridecommandlockouts
\usepackage{cite}
\usepackage{amsmath,amssymb,amsfonts}
\usepackage{algorithmic}
\usepackage{graphicx}
\usepackage{textcomp}
\usepackage{xcolor}
\usepackage{hyperref}
\def\BibTeX{{\rm B\kern-.05em{\sc i\kern-.025em b}\kern-.08em
    T\kern-.1667em\lower.7ex\hbox{E}\kern-.125emX}}
\begin{document}

\title{Variational voxelwise rs-fMRI representation learning: Evaluation of sex, age, and neuropsychiatric signatures\\
{\footnotesize}
\thanks{This work was supported by NIH: R01MH118695}
}

\author{\IEEEauthorblockN{1\textsuperscript{st} Eloy Geenjaar}
\IEEEauthorblockA{\textit{Faculty of EEMCS} \\
\textit{Delft University of Technology}\\
Delft, the Netherlands \\
egeenjaar@gatech.edu}
\and
\IEEEauthorblockN{2\textsuperscript{nd} Tonya White}
\IEEEauthorblockA{\textit{Department of Radiology and Nuclear Medicine} \\
\textit{Erasmus MC}\\
Rotterdam, the Netherlands \\
t.white@erasmusmc.nl}
\and
\IEEEauthorblockN{3\textsuperscript{rd} Vince Calhoun}
\IEEEauthorblockA{\textit{TReNDS center} \\
\textit{GaTech, GSU, \& Emory University} \\
Atlanta, USA \\
vcalhoun@gsu.edu}
}

\maketitle

\begin{abstract}
We propose to apply non-linear representation learning to voxelwise rs-fMRI data. Learning the non-linear representations is done using a variational autoencoder (VAE). The goal is to use a VAE trained on voxelwise rs-fMRI data to perform non-linear dimensionality reduction that retains meaningful information. The retention of information in the model's representations is evaluated using downstream age regression and sex classification tasks for the UK Biobank dataset. The results on these tasks are highly encouraging and a linear regressor trained with the representations of our unsupervised model performs almost as well as a supervised neural network, trained specifically for age regression on the same dataset. Another important quality of our model is that the representations with which age regression and sex classification are performed are the same. This implies that the representations can retain information about both the sex classification and age regression tasks, something a supervised model would have to specifically be trained for with multi-task learning and signifies its potential as a dimensionality reduction method. 

The model is also evaluated with a schizophrenia diagnosis prediction task, to assess its feasibility as a dimensionality reduction method for neuropsychiatric datasets. These results highlight the potential for pre-training on a larger set of individuals who do not have mental illness, to improve the downstream neuropsychiatric task results. The pre-trained model is fine-tuned for a variable number of epochs on a schizophrenia dataset and we find that fine-tuning for $1$ epoch yields the best results. This work therefore not only opens up non-linear dimensionality reduction for voxelwise rs-fMRI data but also shows that pre-training a deep learning model on voxelwise rs-fMRI datasets greatly increases performance even on smaller datasets. It also opens up the ability to look at the distribution of rs-fMRI time series in the latent space of the VAE for heterogeneous neuropsychiatric disorders like schizophrenia in future work. This can be complemented with the generative aspect of the model that allows us to reconstruct points from the model's latent space back into brain space and obtain an improved understanding of the relation that the VAE learns between subjects, timepoints, and a subject's characteristics. 
\end{abstract}


\section{Introduction}

\subsection{Context}
\label{sec:introduction:significance}
Mental disorders are studied through multiple neuroimaging techniques, but resting-state functional MRI (rs-fMRI) has become increasingly important because it allows researchers to image the functional dynamics of a subject's resting brain over time. Progress in the understanding of mental disorders from rs-fMRI data has largely been achieved through linear representation learning techniques, such as independent component analysis (ICA) \cite{mckeown1998independent}. These techniques have opened up the ability to study large-scale functional connectivity differences in complex mental disorders such as autism spectrum disorder (ASD) and schizophrenia, either statically or dynamically \cite{damaraju2014dynamic}. The success of linear representation learning and the increased use of deep learning methods in the field of neuroimaging paves the way towards analyzing these functional differences using deep learning techniques. Findings obtained with deep learning analyses can be complemented with previous research and linear representation learning to move towards individualized predictions \cite{sui2020neuroimaging}, a better understanding of mental disorders, and more effective individualized treatment.

Deep learning methods in fields other than rs-fMRI analysis are often applied to minimally processed data. An important side note here is that these non-fMRI datasets are often also one or two magnitudes larger than many rs-fMRI datasets. The application of deep learning techniques to rs-fMRI data generally involves a dimensionality reduction step after which a supervised neural network is trained to perform classification. Supervised classification is attractive because neural networks gained attention for their outstanding classification performance, most famously on ImageNet \cite{krizhevsky2012imagenet}. The non-linearities that neural networks can model are likely interesting in our understanding of mental disorders as well. A 3-dimensional adaptation of supervised convolutional neural networks was able to obtain robust discriminative neuroimaging biomarkers \cite{abrol2021deep}. 

The methods that are used to perform dimensionality reduction, like independent component analysis (ICA) and principal component analysis (PCA), are unsupervised however. These methods assume that the data is generated through some unseen factors. In case of ICA, these factors are often referred to as intrinsic networks because they are spatially independent and interpretable as separate localized functional networks in the brain. Data-driven methods that are used to find these generative factors, like ICA and PCA, have been extended to restricted Boltzmann machines (RBMs)  \cite{hjelm2014restricted}, which is a precursor to contemporary unsupervised deep learning models like variational autoencoders (VAEs) \cite{kingma2013auto}. VAEs have recently been popularized for non-rs-fMRI data and gained attention due to their interpretable latent space and ability to variationally learn generative factors that fit a certain prior. Previous work evaluates representation learning with VAEs on rs-fMRI data that has first undergone dimensionality reduction \cite{zhao2019variational, matsubara2020deep, kim2020representation, zhang2021spatiotemporal}. These dimensionality reductions may incur overly specific inductive biases and, as a result, limit the expressivity of deep learning methods, especially since neural networks are considered universal function approximators. \cite{hornik1989multilayer}.

\subsection{Problem statement}
\label{sec:introduction:problem}
This work looks at whether unsupervised deep learning methods can learn informative representations from minimally processed voxelwise rs-fMRI data that has not undergone spatial dimensionality reduction(s). Given the high dimensionality of rs-fMRI data, it is useful to be able to reduce the data dimensionality in such a way that we retain information about the variables we want to study. This work, therefore, evaluates the validity and usefulness of the proposed non-linear dimensionality reduction by evaluating the model's latent representations on downstream age regression and sex prediction tasks. Due to its success as an unsupervised representation learning technique, this work uses a variational autoencoder (VAE) \cite{kingma2013auto}. VAEs learns to maximize the lower bound on the marginal likelihood of the training data.

One of the main reasons that this work considers the age regression and sex prediction task as the main downstream tasks, is the availability of a large rs-fMRI dataset that records both demographic factors. In this work, we find that large datasets can be important for representation learning from minimally processed rs-fMRI data. The introduced method is also evaluated on a schizophrenia diagnosis prediction task, with and without first pre-training the model on the larger age regression and sex prediction dataset to study whether pre-training improves downstream performance \cite{mahmood2020whole, erhan2010does}. To evaluate the effect of the dimensionality of the representations on downstream task performance, the age and sex prediction tasks are performed with representations of varying sizes. These results are compared to a linear baseline that performs principal component analysis (PCA) with the same varying number of components.

\section{Related work}
\label{chap:related}
Some interesting prior work has used variational autoencoders to analyze rs-fMRI data. Such work has focused on modeling functional brain networks and ADHD identification \cite{qiang2020deep}, representation learning \cite{kim2020representation}, automatically clustering connectivity patterns \cite{zhao2019variational}, schizophrenia, bipolar disorder and autism spectrum disorder classification \cite{matsubara2020deep}, and spatio-temporal trajectory identification \cite{zhang2021spatiotemporal}. It is important to note however that each of these methods performs spatial and sometimes temporal dimensionality reduction before they use the data as input for their VAE. The non-linear dimensionality reduction on voxelwise rs-fMRI data, that we propose, may however lead to larger gains in terms of meaningful information retention. Using minimally preprocessed data is often seen as the norm in other deep learning fields because it simplifies the data acquisition process and does not reduce any of the potential information in the data.

\section{Variational Autoencoders}
\label{sec:vaes}

VAEs have become an important model architecture for unsupervised representation learning. Variational methods are a way of describing an unknown probability density function that is hard to sample from and/or approximate, byways of parametrizing a simpler probability distribution, such as a Gaussian, in such a way that it can approximate the unknown distribution \cite{mackay2003information}. VAEs provide a neural network-based autoencoder perspective to this problem. Since VAEs are generative models, their goal is to find the underlying latent factors that generate the data.
This ties back to the assumption in methods like independent component analysis (ICA) and principal component analysis (PCA), that the dimensionality of the latent factors is smaller than the original dimensionality of the data. Both of these methods are often used to reduce the dimensionality of rs-fMRI data.

The problem can formally be set as having an rs-fMRI dataset $\{X^{(i)} = \{x^{(i,0)}, …, x^{(i,T)}\}\}_{i=1,…, N}$ with T timepoints and N subjects.
Each $x^{(i,t)}$ is generated from an unknown conditional distribution $p_{\theta}(x^{(i,t)} | z^{(i, t)})$, where $z^{(i, t)}$ is assumed to be a random unseen continuous-valued variable sampled from a prior distribution $p_{\theta}(z)$ \cite{kingma2013auto}.
Both the prior distribution and the conditional distribution are unknown and the integral over the marginal probability of $\bar{x}$: $p_{\theta}(\bar{x}) = \int p_{\theta}(\bar{z}) p_{\theta}(\bar{x} | \bar{z}) \text{d}\bar{z}$ is therefore intractable. Bayesian variational methods can be used to tackle this problem and VAEs have become a common non-linear method to do so \cite{kingma2013auto}.

VAEs approximate the intractable posterior distribution $p_{\theta}(z^{i, t} | x^{i, t})$ using a recognition model, parameterized as a neural network, $q_{\phi}(z^{i, t} | x^{i, t})$ \cite{kingma2013auto}, which can be thought of as an encoder from a coding theory perspective. The conditional distribution $p_{\theta}(x^{i, t} | z^{i, t})$, also parameterized as a neural network, can then be thought of as a decoder. This allows us to perform optimization using the evidence lower bound (ELBO), which is a lower bound on the marginal likelihood of a data point $x^{(i,t)}$ and is explained in detail in the original VAE paper \cite{kingma2013auto}. The objective function, with which the VAE is trained, is the average over all of the data points in a dataset. The loss can then be split into two parts, the first part minimizes the KL-divergence between an apriori selected prior and the distribution that is parameterized by the encoder. Although there is no clear consensus, rs-fMRI data is often seen as or assumed to be normally distributed \cite{laumann2017stability}. The prior ($p_{\theta}(z)$) we thus choose is a diagonal multivariate normal distribution, where each of the dimensions of the normal distribution does not explicitly depend on each other. The second part of then maximizes the log-likelihood of a data point $x^{(i, t)}$ given an estimated latent variable $z^{(i, t)}$. Although each rs-fMRI time point is assumed to be an i.i.d. sample when training the VAE, the temporal relation between the latent variables for a single subject ($z^{(i, 0)}, ..., z^{(i, T)}$) is considered during each prediction task.

\section{Data}
\label{chap:data}
This work uses multiple datasets, one large dataset to do age regression and sex prediction with. The other dataset is used to perform the schizophrenia diagnosis prediction task on.

\subsection{Age and sex dataset}
\label{sec:data:age}
The rs-fMRI data that is used for age and sex prediction are subjects without any diagnoses or self-reported illnesses ($n = 12,314$). These subjects were selected from the 22,392 subjects that were available in the UK Biobank repository on April 7th, 2019 \cite{miller2016multimodal}. The subjects have a mean age of $62.58$ with a standard deviation of $7.41$, $49.6\%$ are female. The youngest subject is 45 years old and the oldest is 80. The scanning parameters are explained in greater depth in the original UK Biobank paper \cite{miller2016multimodal}, however, an important parameter in this work is the repetition time (TR = $0.735$ seconds). With an acquisition time of $6$ minutes, the UK Biobank data acquires a total of 490 time points. The data is minimally preprocessed with the Melodic pipeline \cite{miller2016multimodal} and registered to the MNI EPI template with the help of FMRIB's Linear Image Registration Tool (FLIRT).  The registration is followed by normalization in SPM12, after which it is smoothed with a $6$mm wide FWHM Gaussian kernel. This results in rs-fMRI volumes with a size of $53$ x $63$ x $52$ voxels, and $490$ timepoints per subject.
The size of the volumes and the number of timepoints lead to large memory requirements during training and rs-fMRI data can be noisy. To tackle both problems simultaneously, we use a piecewise aggregate approximation (PAA) to reduce the noise and memory consumption, while still keeping the trend of the time series. PAA takes the average over points in consecutive windows with a certain window size. For the UK Biobank dataset, the window size is set to $15$, which is equivalent to taking the average over a period of about $11$ seconds. This reduces the number of time points to $33$.

\subsection{FBIRN}
\label{sec:data:schizophrenia}
The dataset that is used to evaluate whether meaningful information about a mental illness is retained in the representations is the FBIRN study \cite{keator2016function}. The dataset is processed using the NeuroMark preprocessing pipeline \cite{du2020neuromark} to obtain rs-fMRI volumes with a size of $53$ x $63$ x $52$ voxels. The repetition time for FBIRN is $2$ seconds. To stay in line with the temporal preprocessing that is done for the UK Biobank dataset, we apply PAA to FBIRN as well, but to account for the different repetition times, the window size that is used is $5$. This corresponds to a period of $10$ seconds.

\section{Methodology}
\label{chapter:method}
The downstream task performances will be compared to a linear baseline method. There is not much previous work that looks at voxelwise rs-fMRI representation learning, although other work focuses on supervised sex classification and age regression on the same dataset used in this work \cite{anees2021}. Supervised methods are however not a comparable baseline, although it is insightful as a bound to strive towards. A linear dimensionality reduction method that is in some sense comparable to VAE in terms of what components it tends to learn is principal component analysis (PCA) \cite{rolinek2019variational}. Given that PCA and a VAE are comparable and that there are online versions of PCA \cite{artac2002incremental} that do not have large memory requirements, the linear baseline used in this work is IncrementalPCA \cite{artac2002incremental, scikit-learn}.

Since one dataset is significantly smaller and is prone to overfitting, this work also evaluates whether a model that is pre-trained on a large dataset like UK Biobank (n=$12,314$) can be fine-tuned on FBIRN (n=$325$) to improve representations for schizophrenia diagnosis prediction. We compare pre-trained models that have been fine-tuned for a variable number of epochs on the FBIRN dataset.

\subsection{Model architecture}
\label{sec:method:architecture}
The architecture of the model is based on a ResNet \cite{he2016deep}. Each residual block in the encoder and decoder has a skip connection. These skip connections allow the network to learn longer dependencies and have been used in VAEs before \cite{kingma2016improving} to improve their variational inference. The activations that are used in the network are exponential linear units (ELUs) \cite{clevert2015fast}. Instead of batch normalization, the network uses weight normalization \cite{salimans2016weight}. Both the activation function and weight normalization are considered best practices when training VAEs \cite{salimans2016weight}. Batch normalization may lead to drift during inference in a VAE which can cause unstable results. 

The encoder consists of five residual blocks, with $16$, $32$, $64$, $128$, and $256$ output channels for each block, respectively. These blocks all downscale their original inputs by two until the last residual block produces an output feature map of $256$ x $2$ x $2$ x $2$, which is flattened to $2048$ features. These features are then used to learn the mean and variance of the multivariate Gaussian in the latent space from which latent variable $z$ is sampled. The layer computes the square root of the natural logarithm of the variance instead of the standard deviation to increase the stability of training the network and to make sure variations due to gradient updates have a smaller effect on standard deviations near zero. This allows the network to model the standard deviations near zero more accurately. The first layer in the decoder is a linear layer that maps latent variable $z$ to $2048$ features, which are then reshaped to a $256$ x $2$ x $2$ x $2$ feature map. We use trilinear interpolations on the feature maps with a scale of two to double the size. The rest of the decoder consists of five residual blocks, where each residual block is preceded by a trilinear interpolation layer. The final layer is a $1$ x $1$ x $1$ transpose convolutional layer, without an activation function. In earlier iterations of this work we tried to use a sigmoid activation on the last layer, this leads to instability, because the combination of the mean squared error (MSE) as a loss function and a sigmoid leads to a non-convex objective function. Each of the layers is initialized according to work that proposed ResNets \cite{he2016deep}, this initialization is also used in the original ELU paper \cite{clevert2015fast}. The VAE is trained for 100 epochs using the ADAM optimizer \cite{kingma2014adam} with a learning rate of $5E-4$. Before the input data is used it is first rescaled to be between [0, 1], values below $0.05$ are then thresholded to remove possible background noise.

\subsection{Regression and classification}
After training the VAE, there are multiple ways to evaluate what information is contained in the representations ($z^{(i, 0)}, …, z^{(i, T)})$. To evaluate whether the temporal information improves classification and regression with simple machine learning classifiers, these classifiers are trained with a subject's latent temporal average $z^{(i, \mu)}$ and also with a subject's concatenated latent time series. The machine learning classifiers that are used in this work are a support vector machine (SVM) and a k-nearest neighbor classifier (kNN) for the classification tasks and a support vector regression (SVR) and k-neighbor regressor (kNR) for the age regression task. These classifiers give us insight into the linear separability of the representations (SVM) or how well they are clustered (kNN). To take the temporal information between the representations into account more specifically, we also train a long-short term memory (LSTM) \cite{hochreiter1997long} on the full latent time series. The LSTM is either trained with a mean squared error (MSE) for the regression task or a binary cross-entropy (BCE) loss for the classification task. The hidden size for the hidden states in the LSTM ($h^{(0)}, …, h^{(T)}$) are twice the size of the input representations, and all of the hidden states in the LSTM are concatenated together to form a feature vector that is then mapped to a prediction using a linear layer. This allows the model to learn from the hidden state at each timestep more directly. The size of that feature vector can be quite big, so we apply dropout to that last linear layer. This is a common technique to counter overfitting and promote a more robust prediction model \cite{srivastava2014dropout}.

\subsection{Evaluation measures}
To be able to compare the results obtained using each of the methods, we use multiple evaluation measures. The first measure is used for the classification tasks and computes the area under the receiver operating characteristic (ROC-AUC), which is a more complete way of comparing binary classifiers. To evaluate the regression task, we use three measures, the first is the mean average error (MAE) which is the L1-norm between the predicted age and the correct age. The second measure is the R2-score, which is also referred to as the coefficient of determination. The last measure is the Pearson product-moment correlation between the predicted ages and the true ages.

\section{Experiments}
\label{sec:experiments}
The code for the VAE was implemented by the authors in PyTorch \cite{NEURIPS2019_9015}, training was performed with Catalyst \cite{catalyst} and TorchIO \cite{perez-garcia_torchio_2020}, and the regression and classification pipelines were implemented using RAPIDS-AI \cite{rapidsai}, scikit-learn \cite{scikit-learn}, and NumPy \cite{van2011numpy}. To minimize costly transfers between the CPU and the GPU, most of the classifications were done using RAPIDS-AI \cite{rapidsai} , to make sure the computed representations could be kept in GPU memory without any copies or transfers from or to the CPU. All of the experiments were performed on an NVIDIA DGX-1 V100. Due to time restrictions, the UK Biobank experiments could only be performed on one train and test split, because each VAE epoch takes around 45 minutes on a single GPU. Training on multiple training and test folds is essential for the schizophrenia diagnosis prediction task because the variance between predictions can be large, especially for deep learning models. The schizophrenia results are thus trained over 5-folds, where one fold is used as the held-out set and the other four folds are used as a training and validation set. To make sure the model does not overfit, we use an early stopping criterion that stops the model if its loss objective has not improved on the validation set for 20 epochs. Further, instead of taking the $z$ that is sampled from the distribution that the encoder outputs, we use the mean of that distribution. This is to reduce the stochasticity during inference. The reason we do not use the standard deviation is that it did not improve our preliminary results. 

To determine the effect that the size of the representations has on the performance of the age regression and sex classification tasks, the model is trained with multiple latent dimensionalities, specifically: $64$, $128$, $256$, and $512$. These tests are done on the UK Biobank dataset because we noticed that training on a larger dataset is more stable. 

We tested whether initializing a model for the schizophrenia classification task with a pre-trained model on UK Biobank improves the results on that task. The number of latent dimensions that are used is $256$. The fine-tuning is evaluated for $0$, $1$, $2$, $5$, $10$,  $50$, and $100$ epochs and compared to a non-pre-trained model, and the baseline.

The baseline for this work is IncrementalPCA, which is implemented in scikit-learn \cite{scikit-learn}. Similar to the VAE, the principal components are obtained for each volume in a subject’s time series independently. The components are whitened and temporally averaged. They are then used as input to the downstream classifiers/regressors.

\section{Results}
\label{sec:results}
The age and sex downstream tasks are evaluated for multiple latent dimensionalities and compared with a baseline PCA that has the same number of components. The classification methods are referred to as SVM/SVR and kNN/kNR when the representations for each timestep are concatenated to create a single feature vector ($z^{(i, 0)}$, ..., $z^{(i, t)}$) for each subject. Classifiers mSVM/mSVR and mkNN/mkNR take the average representation over the timesteps as input for each subject. The $128$-dimensional VAE did not converge well which led to worse results for that specific model. It is also important to recognize that the representations used to perform age regression are the same as the ones used to do sex classification. The model is thus able to retain information about both the sex classification and age regression tasks, something a supervised model would have to specifically be trained for with multi-task learning. 

\subsection{Age regression results}
\begin{figure*}[!htb]
\begin{center}
    \includegraphics[width=0.9\linewidth]{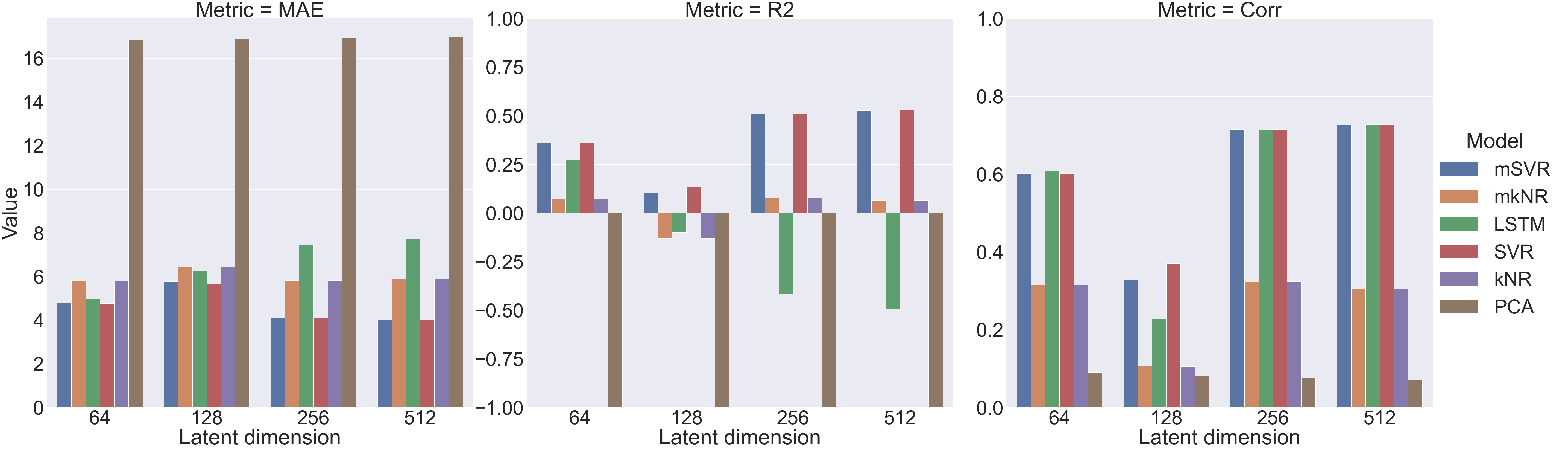}
\end{center}
\caption{This figure portrays the results for the downstream age regression task, evaluated using three metrics, MAE (mean absolute error), the R2 score, and the correlation (Corr) between the predictions and the ground truth values. Each bar plot shows all of the models at the $4$ different latent dimensionalities: $64$, $128$, $256$, $512$ on the x-axis. The run with a $128$-dimensional VAE did not converge well, an anomaly, which leads to worse results but, a VAE-SVR performs best overall. Note that for the MAE lower is better, and for the R2 score and correlation higher is better. The baseline performs significantly worse on all three of the metrics. Further, the VAE-LSTM performs relatively well on the correlation metrics but significantly underperforms on the MAE and the R2 score. The R2-scores for the baseline are cut out of this figure because they are too low and displaying them would reduce the interpretability of the figure, those values are roughly $-5$.}
\label{fig:age-results}
\end{figure*}

The age regression task is evaluated using three measures, the mean absolute error (MAE) is seen as most important in this work, the results for the task are shown in Figure \ref{fig:age-results}. All of the VAE models, even the non-converging $128$-dimensional VAE, outperforms the baseline PCA method. The best performing model is the $512$-dimensional VAE-SVR with an MAE of $4.014$ years, an R2 score of $0.5288$, and a correlation between the predicted and ground truth ages of $0.727$. The general trend for the number of latent dimensions is that more latent dimensions improve the downstream performance on the age regression task. The difference between the $256$-dimensional VAE and the $512$-dimensional VAE is however significantly smaller than between the two smaller latent dimensionalities. Another interesting result is that the SVR and mSVR always outperform the kNR and mkNR, which suggests that the latent space is linearly separable for the age regression task, as opposed to being clustered based on age. Furthermore, the SVR and mSVR also outperform the LSTM. The LSTM is a sequential method and is therefore good at representing temporal data, it may struggle with data where temporal relations do not aid in regression improvements. The LSTM does perform well on the correlation between the predicted and ground truth ages. The lower performance of the LSTM and the near equivalent performance of the mSVR and SVR may imply that there is no (linear) temporal relationship between age and the VAE's representations.

Comparatively, previous work \cite{anees2021}  reports that on the same dataset, a voxelwise supervised deep learning model achieves an MAE of $3.54$ years, an R2 score of $0.65$, and a correlation of $0.82$. It is important to note that the VAE model in this work received no supervised signal to model the features necessary to achieve its downstream age regression results. Further, the same previous work \cite{anees2021} finds that using the ICA time series achieves an MAE of $4.66$, highlighting the benefits of using our proposed dimensionality reduction because it outperforms the ICA time series. Further, they find that taking the mean over the temporal dimension before using the input for the classification task improves performance, which seems to be the same in this work.

\subsection{Sex classification results}

\begin{figure}[htb]
\begin{center}
    \includegraphics[width=0.8\linewidth]{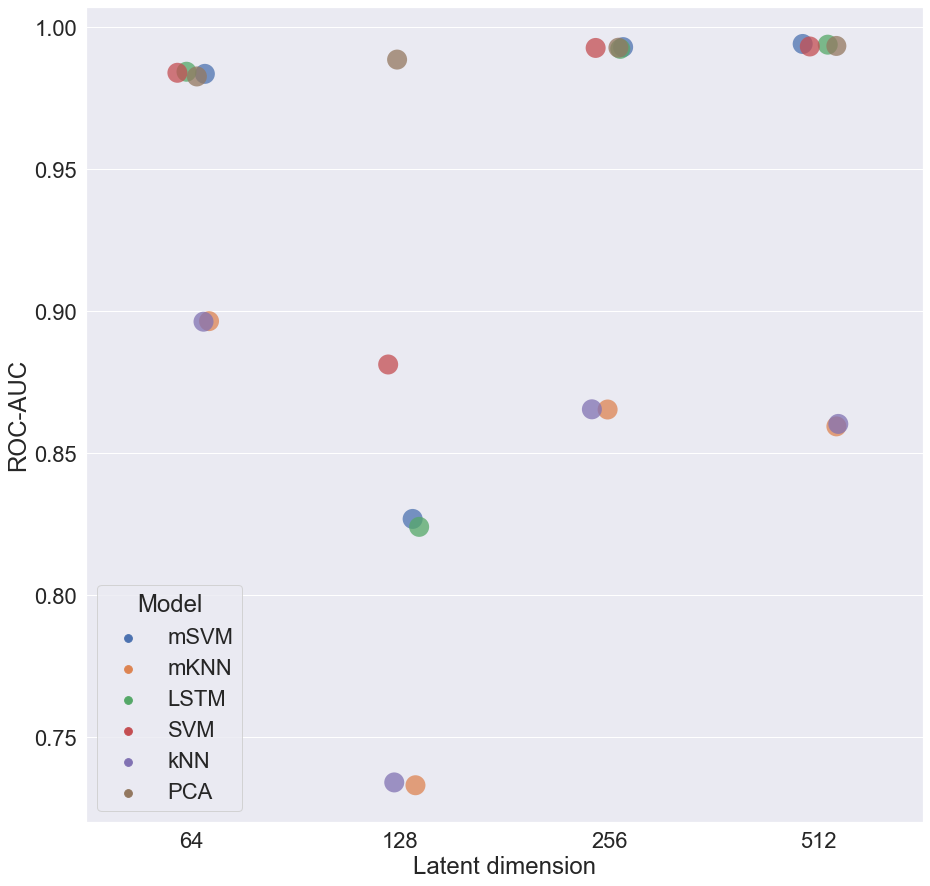}
\end{center}
\caption{The results for the downstream sex classification task, the area under the curve of the receiver operating characteristic (ROC-AUC) is shown on the y-axis. The $4$ different latent dimensionalities are shown on the x-axis: $64$, $128$, $256$, and $512$. The run with a $128$-dimensional VAE did not converge well. The models perform similarly to the baseline, although the VAE-mSVM and VAE-SVM perform slightly better. The highest ROC-AUC is achieved using a $512$-dimensional VAE-mSVM: $0.994$}
\label{fig:sex-results}
\end{figure}

The results in Figure \ref{fig:sex-results} show that the age classification task for this dataset is fairly trivial and the baseline performs only slightly worse than the VAE-SVM and VAE-mSVM. The baseline outperforms the $128$-dimensional VAE because it did not converge well. 

All of the models improve with increasing latent dimensionality, although the model only slightly improves with more than $256$ dimensions. Interestingly, in contrast to the age regression results, the LSTM performs only slightly worse than the mSVM and SVM. The mkNN and kNN are always outperformed by the mSVM, SVM, and LSTM. The best performing model is the $512$-dimensional VAE-mSVM with an ROC-AUC of $0.994$. In general, except for the $128$-dimensional VAE, the mSVM outperforms the SVM, which suggests that the temporal information for each subject does not help with the linear separability of sex in the latent space. The differences between the SVM and the mSVM are likely insignificant in terms of determining which one is better, but the SVM does not seem to have any performance gain over the mSVM.

\subsection{Schizophrenia classification results}
\begin{figure}[htb]
\begin{center}
    \includegraphics[width=0.8\linewidth]{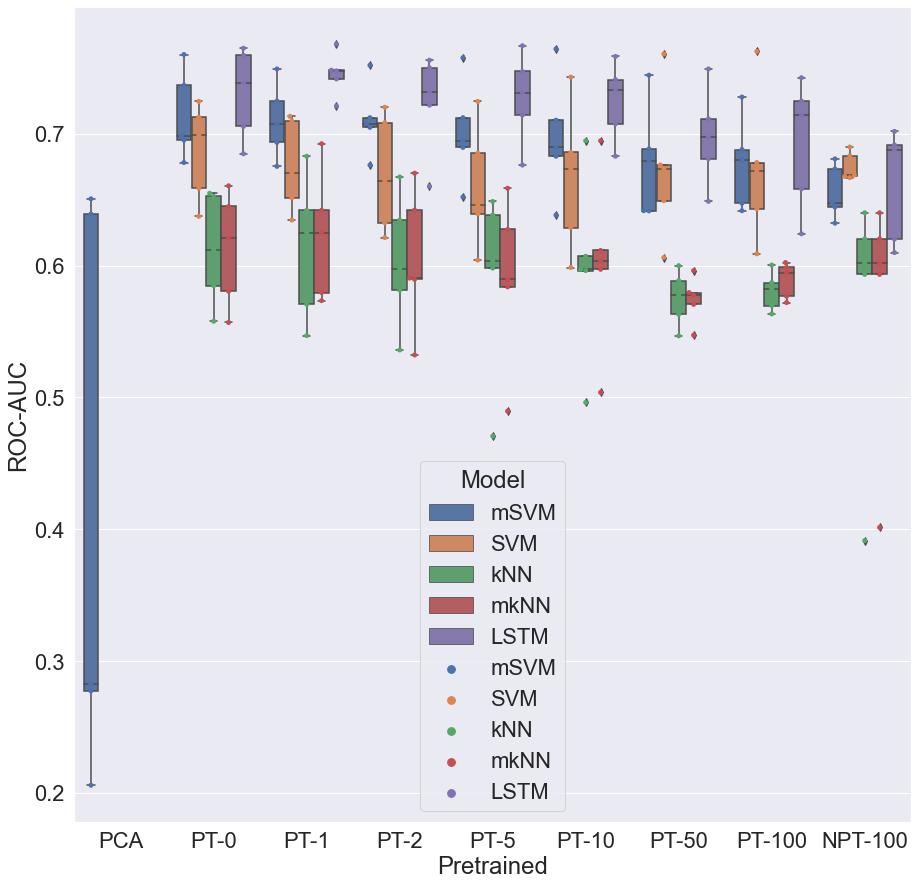}
\end{center}
\caption{The results for the downstream schizophrenia classification task, the area under the curve of the receiver operating characteristic (ROC-AUC) is shown on the y-axis. The baseline (PCA), pre-trained (PT), and non-pre-trained (NPT) models are shown on the x-axis. Pre-training seems to generally improve the ROC-AUC, especially for the VAE-LSTM. The longer the model is fine-tuned on FBIRN, the more it seems to overfit, which leads to worse results. Fine-tuning for one epoch seems to be optimal.}
\label{fig:sz-results}
\end{figure}

The downstream schizophrenia diagnosis classification task is evaluated for a pre-trained (PT) and a non-pre-trained model (NPT). The pre-trained model is the same for every run but is fine-tuned on the schizophrenia diagnosis prediction task for a variable number of epochs. It turns out that fine-tuning the pre-trained model for $1$ epoch in combination with the downstream LSTM yields the best (an average ROC-AUC of $0.7452$) and least variable results over multiple folds. It is clear from Figure \ref{fig:sz-results} that, especially in combination with the LSTM, fine-tuning with more epochs leads to worse downstream performance. This is likely because the model starts to overfit on FBIRN, which is likely due it being a much smaller dataset. It is also notable that the pre-trained model that is fine-tuned for $100$ epochs on FBIRN in combination with the LSTM still performs better than the model that is only trained on FBIRN for $100$ epochs. Further, the fact that the LSTM performs the best indicates that the temporal information in the representations is related to the schizophrenia diagnosis of the subjects. It may also imply that the information about schizophrenia diagnosis is non-linearly entangled in the representations. The mSVM does generally outperform the SVM, which may further suggest that the temporal information in the representations is at least not linearly related to a schizophrenia diagnosis. We also find that the kNN-based classifiers are consistently outperformed by the other classifiers and that the baseline ranks lowest in terms of performance.

To compare this to other work, we look at Whole MILC \cite{mahmood2020whole}, which uses a novel form of pre-training. Their best supervised classification model achieves an average ROC-AUC on FBIRN of roughly $0.79$-$0.80$, which is not that much higher than what we have achieved with our unsupervised representations. Note that they utilize data that has undergone dimensionality reduction using ICA with a spatially constrained prior \cite{fu2019altered}. Our pre-training results are also similar in that we both find an improvement in downstream task performance after pre-training, although they do not use an unsupervised autoencoder as the primary model that they propose in their work.

\subsection{Sex-based group differences voxel space}
\label{sec:sex-differences}
\begin{figure}[htb]
\begin{center}
    \includegraphics[width=0.8\linewidth]{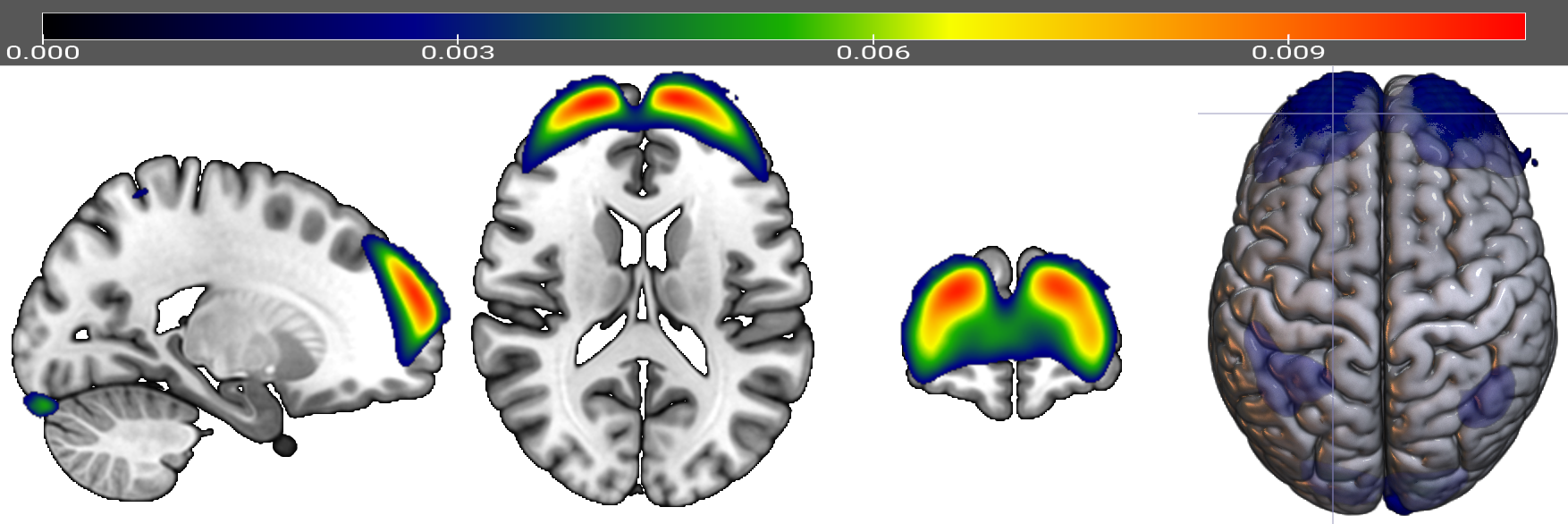}
\end{center}
\caption{The brain differences in females compared to males, calculated by subtracting the reconstructed average latent representation for males from the reconstructed average latent representation for females. The volume is then thresholded at the highest 80th quantile absolute value. The visualization shows significantly higher activation in the prefrontal cortex and some small increased activation in the parietal lobules and between the cerebellum and occipital lobe.}
\label{fig:sex-diff}
\end{figure}

A VAE is a generative model and is thus capable of reconstructing locations from the latent space. To visualize the differences in the VAE models between males and females in its latent space, the average representation for both groups is decoded back into the voxel space and the reconstruction for males is subtracted from the reconstruction for females. The resulting volume is then thresholded at the highest 80th quantile absolute value and the differences are shown in Figure \ref{fig:sex-diff}. The decoded group-wise differences show that women on average have increased rs-fMRI activation in a large area of the prefrontal cortex, which has been reported in the literature before \cite{hill2014gender}. There also appears to be some increased average activity in the left and right inferior parietal lobules. The activity in the inferior parietal lobules and between the occipital lobe and the cerebellum looks more like noise and does not persist when higher thresholds are used.

\section{Conclusion}
\label{sec:conclusion}
This work investigated whether unsupervised deep learning techniques, more specifically a VAE, can learn robust representations that can be used in downstream neuroimaging tasks from minimally preprocessed voxelwise rs-fMRI data. The representations learned by a VAE  were evaluated on multiple downstream tasks and for multiple different latent dimensionalities on the UK Biobank dataset. The larger the dimensionality, the better the performance. It turned out that the sex classification task was rather trivial, but the difference between the baseline and our model became clear on the age regression task. The model was also evaluated on FBIRN, a schizophrenia dataset, both without and with pre-training on UK Biobank first. The optimal number of fine-tuning epochs on FBIRN turned out to be $1$ epoch. Our results open up more work into non-linear dimensionality reduction of voxelwise rs-fMRI data with unsupervised deep learning methods while retaining meaningful information.

\section{Discussion}
\label{sec:discussion}
The results in this work show that there is great potential for voxelwise rs-fMRI representation learning with a VAE. The representations that are learned by the VAE contain information that allows linear classifiers and regressors to predict the sex and age of a subject with high precision. The model also performed well on a downstream schizophrenia diagnosis prediction task. The results on the latter especially opened up more work into the pre-training of unsupervised models like the one proposed in this work. The pre-trained model is first trained on a larger dataset before fine-tuning it on neuropsychiatric datasets for a few epochs and using the final model to encode low-dimensional representations. Even without fine-tuning our pre-trained model outperformed a model that was trained only on FBIRN.

Important future considerations include the need for a more efficient solution in terms of computational runtime. An important bottleneck during this project was loading the large data files onto the GPU. Improving data movement and transfer will make experimentation more feasible and may also address the shortcomings that smaller batch sizes may have on the training dynamics of a neural network. If training times can be reduced, it is also possible to train on even larger datasets and perform hyperparameter tuning on a large scale. Larger datasets will likely improve the results on all of the downstream tasks, similar to the improvements pre-training had on the schizophrenia diagnosis prediction task's performance. Smaller datasets are more likely to lead to overfitting, especially since voxelwise rs-fMRI data is highly noisy. Not only should future work look at larger datasets, but also move towards models that more efficiently use the information available in (smaller) datasets. Models that incorporate inductive biases and/or forms of regularization are required to move towards meaningful non-linear representations for mental illnesses from voxelwise rs-fMRI data.

It is also relevant to look at the manifold that our model has learned, especially in terms of the schizophrenia diagnosis prediction task. The inherent generative quality of this model supplies us with a wide range of possible interpretability analyses, because representations can be mapped back into brain space using the model's decoder. One simple example of a generative interpretability analysis was provided in Section \ref{sec:sex-differences}. More complex, temporal, analyses should be designed for complex neuropsychiatric disorders like schizophrenia. Complex neuropsychiatric disorders could also benefit from studying a single subject representation, that combines the temporal information in a subject's time series into a single vector.

\end{document}